\documentclass[10pt,twocolumn,letterpaper]{article}

\usepackage{cvpr}              

\usepackage{graphicx}
\usepackage{amsmath}
\usepackage{amssymb}
\usepackage{booktabs}
\usepackage{multirow}
\usepackage{enumitem}

\usepackage[pagebackref,breaklinks,colorlinks]{hyperref}

\usepackage[capitalize]{cleveref}
\crefname{section}{Sec.}{Secs.}
\Crefname{section}{Section}{Sections}
\Crefname{table}{Table}{Tables}
\crefname{table}{Tab.}{Tabs.}

\def\cvprPaperID{2921} 
\def\confName{CVPR}
\def\confYear{2023}

\makeatletter
\def\thanks#1{\protected@xdef\@thanks{\@thanks
        \protect\footnotetext{#1}}}
\makeatother

\begin{document}

\title{A Benchmark of Video-Based Clothes-Changing Person Re-Identification}

\author{Likai Wang\thanks{\textsuperscript{\rm *}Equal contribution.}\textsuperscript{\rm *1}, Xiangqun Zhang\textsuperscript{\rm *1}, Ruize Han\textsuperscript{\rm 1}, Jialin Yang\textsuperscript{\rm 1}, Xiaoyu Li\textsuperscript{\rm 1}, Wei Feng\textsuperscript{\rm 1}, Song Wang\textsuperscript{\rm 2}\\
\textsuperscript{\rm 1}Tianjin University, China \quad
\textsuperscript{\rm 2}University of South Carolina Columbia, USA\\
{\tt\small \{kkww, clzxq, han\_ruize, abdullahyang, xiaoyuli\}@tju.edu.cn, wfeng@ieee.org, songwang@cec.sc.edu}
}
\maketitle

\begin{abstract}
Person re-identification (Re-ID) is a classical computer vision task and has achieved great progress so far.
Recently, long-term Re-ID with clothes-changing has attracted increasing attention. However, existing methods mainly focus on image-based setting, where richer temporal information is overlooked.
In this paper, we focus on the relatively new yet practical problem of clothes-changing video-based person re-identification (CCVReID), which is less studied. We systematically study this problem by simultaneously considering the challenge of the clothes inconsistency issue and the temporal information contained in the video sequence for the person Re-ID problem. Based on this, we develop a two-branch confidence-aware re-ranking framework for handling the CCVReID problem. The proposed framework integrates two branches that consider both the classical appearance features and cloth-free gait features through a confidence-guided re-ranking strategy. This method provides the baseline method for further studies. Also, we build two new benchmark datasets for CCVReID problem, including a large-scale synthetic video dataset and a real-world one, both containing human sequences with various clothing changes.
We will release the benchmark and code in this work to the public.
\end{abstract}

\section{Introduction}
\label{sec:intro}

\begin{figure}
  \centering
  \begin{subfigure}{0.4\linewidth}
    \includegraphics[width=1.\linewidth]{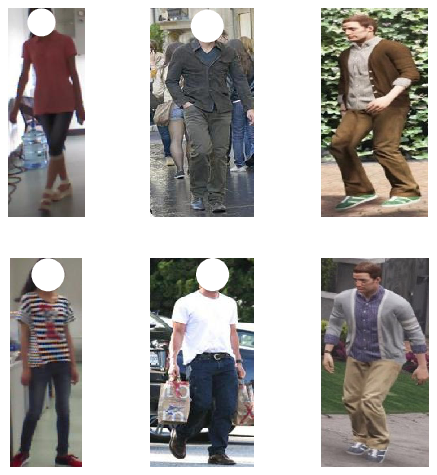}
    \caption{}
    \label{fig:1a}
  \end{subfigure}
  \hfill
  \begin{subfigure}{0.52\linewidth}
    \includegraphics[width=1.\linewidth]{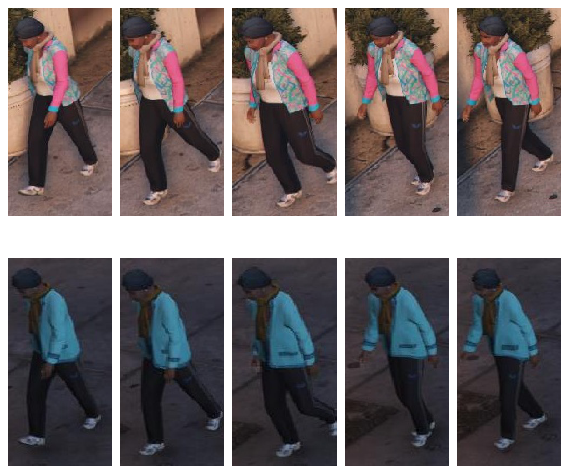}
    \caption{}
    \label{fig:1b}
  \end{subfigure}
  \caption{Examples in clothes-changing benchmark datasets.
  (a) Image-based datasets PRCC~\cite{yang2019person}, Celeb-reID~\cite{huang2019beyond},  and VC-Clothes~\cite{wan2020person} (from left to right). Each column shows a pair of samples of the same person in different clothes. It is can be seen that cased by clothes-changing, appearance information show great bias in the single image.
  (b) Video-based dataset SCCVReID collected in this paper. Each row shows a video sequence of the same person in different clothes. It is can be seen that the clothes-irrelevant information, \textit{e.g.}, human gait, can be clearly obtained.}
  \label{fig:moti}
  \vspace{-10pt}
\end{figure}

The past few years have witnessed that person re-identification (Re-ID) has become a very popular topic in the computer vision community, because of its significance in many real-world applications, such as video surveillance, unmanned supermarket, \textit{etc}.
The classical person Re-ID task~\cite{zheng2015scalable,li2014deepreid,zheng2017unlabeled} focuses on the image-based data, in which the clothes of a person also keep unchanged in the dataset.
This may limit the information utilization from the video sequence, which is easy to obtain in many real-world application.
Also, for the long-period applications, \textit{e.g.}, history criminal retrieval, the assumption of unchanged clothes is not reasonable. 
This way, two categories of person Re-ID task have been proposed in recent years, \textit{i.e.}, the video-based person Re-ID~\cite{mclaughlin2016recurrent,liu2019spatial,eom2021video} and the clothes-changing person Re-ID~\cite{yang2019person,huang2019beyond, qian2020long, yu2020cocas}.
The former focuses on utilizing the temporal information from the video sequence, and the later aims to extract the clothes-independent appearance features for person identification.

In this paper, we are more interested in studying the video-based clothes-changing person Re-ID problem. 
This is because the video-based Re-ID is complementary to the clothes-changing Re-ID.
Specifically, the (image-based) clothes-changing person Re-ID is very challenging given the very limited information from a single human image, which is dominated by the appearance of the clothes (as seen in~\cref{fig:1a}). 
If a video sequence is given, we can obtain more information not related to the clothes, \textit{e.g.}, the human gait (as seen in~\cref{fig:1b}).
However, the video-based clothes-changing person Re-ID problem has not been studied widely.
For this problem, only one dataset namely CCVID~\cite{gu2022clothes} is publicly available, which is actually built based on a gait recognition dataset~\cite{zhang2019gait}.
Although clothes-changing is considered, all samples in the dataset are taken from the same camera view, and all identities follows the same designated route.
Thus, it can not meet the request of person Re-ID problem, \textit{e.g.}, the various scenes, varied perspectives, abundant samples. 

In this work, we propose to build a benchmark to systematically study the Clothes-Changing Video-based Re-ID (CCVReID) problem.
For this purpose, the first step is to collect the applicable datasets.
Considering the convenience of dataset collection and the huge success in cross-domain person Re-ID~\cite{sun2019dissecting,wan2020person}, we first build a large-scale synthetic dataset for CCVReID.
We use a game namely \textit{Grand Theft Auto V} for data collection, whose character modeling is very realistic.
We record the pedestrians for 48 hours in total, from which we obtained contains 9,620 sequences from 333 identities and each identity has 2-37 suits of clothes, with an average of 7. The number of sequences, identities and suits for each person are much larger than previous datasets.
Besides, we also build a real dataset including the human sequences with various clothe changes.

We also propose a new baseline for the CCVReID problem. Our basic idea is to take advantage of the clothes-independent appearance feature from each RGB image and the temporal-aware gait feature from the video sequence. Specifically, we first use a video-based Re-ID method~\cite{bai2022salient} and a gait recognition method~\cite{chao2021gaitset} to extract the appearance and gait features, respectively.
We then conduct the inference stage of the Re-ID task that using a query to get the candidate ranking from the gallery, using the appearance and gait representations, respectively. We propose a two-branch ranking fusion framework, containing a candidate relation graph, to combine the ranking lists from these two representations. We also propose a confidence-aware re-weighting strategy to estimate the representation certainty for balancing the two branches.
We conduct the experiments on an existing dataset and two new datasets collected in this work. The results verify the effectiveness of our method, which outperforms the methods using the appearance and gait separately, with an obvious margin. 

We summarize the main contributions in this work:
\begin{itemize}[itemsep=2pt,topsep=0pt,parsep=0pt]
\item  We systematically study a relatively new and practical problem of Clothes-Changing Video-based Re-ID (CCVReID). For this problem, we simultaneously consider the clothes-inconsistent appearance and the temporal information contained in the sequence to complement each other for person Re-ID, which is overlooked previously.
\item We develop a preliminary framework for handling the CCVReID problem. The proposed framework integrates two branches that focus on the classical appearance and cloth-free biological features (specifically the gait feature) through a confidence-balanced re-ranking strategy. This method can provide the baseline for further studies.
\item We build two new benchmark datasets for the proposed problem, \textit{i.e.}, a large-scale synthetic video dataset and a real-world one, both including the human sequences with various clothe changes.
\end{itemize}

\begin{figure*}[t]
  \centering
   \includegraphics[width=0.9\linewidth]{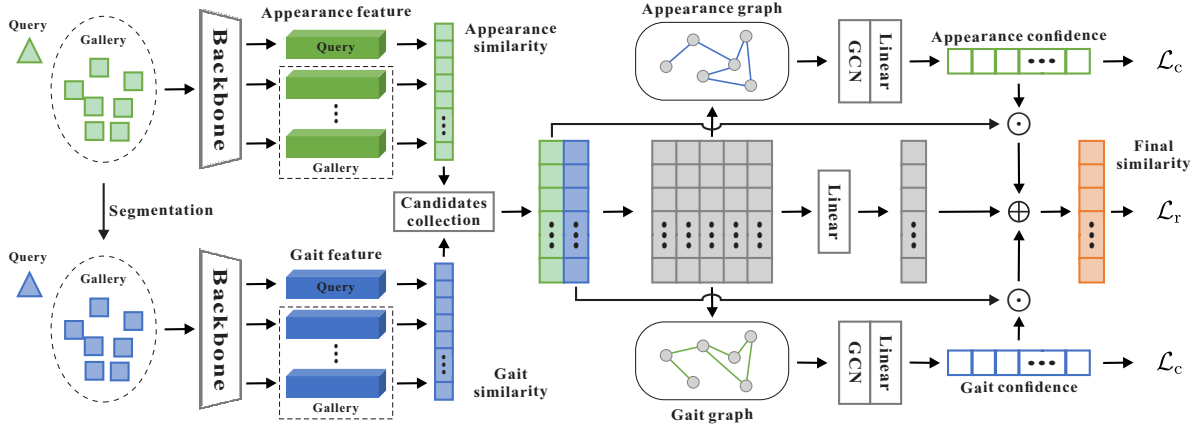}
   \caption{Framework of the proposed method for CCVReID. Given a query and several gallery samples, we first extract their appearance and gait features, and calculate the initial similarity of each gallery to the query.
Based on the obtained appearance and gait similarity, we collect a certain number of candidates, build appearance and gait graph, and use GCN to estimate confidence for each original appearance and gait similarity.
Finally, we use the confidence to re-weight the initial similarities and fuse all information for final re-ranking.}
   \label{fig:method}
   \vspace{-10pt}
\end{figure*}

\section{Related Work}
\label{sec:rel}
\noindent\textbf{Video-based person Re-ID.} Video-based person Re-ID aims to extract spatial-temporal features from consecutive frame sequences. To this end, some of existing methods lift image-based Re-ID methods by aggregating multi-frame features through RNNs~\cite{mclaughlin2016recurrent,liu2019spatial}, mean/max pooling~\cite{zheng2016mars,chung2017two}, and temporal attention~\cite{fu2019sta,chen2018video}, \textit{etc}. Other methods perform concurrent spatial-temporal information modeling via 3D convolution~\cite{gu2020appearance,li2019multi} or graph convolution~\cite{chen2022keypoint}. Despite the fact that a number of works have been done for video-based Re-ID, most of them focus on clothes consistent setting only, which are not applicable to long-term application scenarios.\\
\textbf{Clothes-changing person Re-ID.}
Image-based clothes-changing person Re-ID has been widely studied in the literature. Several public datasets, \textit{e.g.}, PRCC~\cite{yang2019person}, Celeb-reID~\cite{huang2019beyond}, LTCC~\cite{qian2020long}, VC-Clothes\&Real28~\cite{wan2020person}, and COCAS~\cite{yu2020cocas}, have been collected successively to support this task.
Based on these datasets, \cite{huang2019beyond} proposed to use vector-neuron capsules instead of the traditional scalar neurons, to perceive cloth changes of the same person. \cite{li2021learning} proposed a Clothing Agnostic Shape Extraction Network (CASE-Net) to shape-based feature representation via adversarial learning and feature disentanglement.
Besides, other works attempted to leverage contour sketch~\cite{yang2019person}, silhouettes~\cite{hong2021fine,jin2022cloth}, face~\cite{wan2020person}, skeletons~\cite{qian2020long}, 3D shape~\cite{chen2021learning}, or radio signals~\cite{fan2020learning} to capture clothes-irrelevant features.
Despite the achievement of image-based methods, they are susceptible to the quality of person images, \textit{i.e.}, they are less tolerant to noise due to the limited information contained in a single frame.

Recently, some works~\cite{zhang2018long,zhang2020learning,gu2022clothes} have extended clothes-changing person Re-ID to video-based data. 
\cite{zhang2018long} collected a Motion-ReID dataset with clothes-changing, and developed a FIne moTion encoDing (FITD) model based on true motion cues from videos. 
\cite{zhang2020learning} collected a Cloth-Varying vIDeo re-ID (CVID-reID) dataset, which contains video tracklets of celebrities posted on the Internet, and proposed to learn hybrid feature representation from image sequences and skeleton sequences. 
Furthermore, \cite{gu2022clothes} constructed a Clothes-Changing Video person re-ID (CCVID) dataset from a gait recognition dataset FVG~\cite{zhang2019gait}, and proposed a Clothes-based Adversarial Loss (CAL) to mine clothes-irrelevant features from the original RGB images.
In the aforementioned datasets, only CCVID~\cite{gu2022clothes} dataset is publicly available. However, since the original FVG~\cite{zhang2019gait} dataset contains only one view, and the field of view contains only one person that follows designated routes, the CCVID dataset built on that is not suitable for real-world scenarios, where there are usually multiple people who walk in arbitrary directions.
For supporting the CCVReID task, we build two new benchmark datasets of surveillance scenarios in this paper.\\
\textbf{Person Re-ID with gait.} 
Affected by lighting, clothing, view, and other factors, the same person’s appearance features may change a lot, which brings challenges to the Re-ID task. To solve that, \cite{zhang2020learning,lu2022long} attempted to introduce gait to achieve complementarity of appearance and gait features. \cite{jin2022cloth} performed gait prediction and regularization from a single image, which used gait to drive the main Re-ID model to learn cloth-independent features. 
However, such methods only consider the fusion of body and face information in a single sample. In the proposed method, in addition to considering this, we also introduce the contextual information among neighbor samples.\\
\textbf{Synthetic datasets.}
Due to the difficulty of building large-scale datasets in the real world and the cost of extensive manual annotation, synthetic datasets are gaining increasing attention in many computer vision tasks, including pose estimation~\cite{fabbri2018learning,hoffmann2019learning}, tracking~\cite{fabbri2018learning}, action recognition~\cite{varol2021synthetic}, and semantic segmentation~\cite{ros2016synthia,mccormac2017scenenet}, \textit{etc}.
For person Re-ID, \cite{bak2018domain,sun2019dissecting,wan2020person}     have introduced synthetic data. Although the PersonX~\cite{sun2019dissecting} and VC-Clothes~\cite{wan2020person} dataset involve clothes changing, they are all image-based. 
To promote the researches on CCVReID, we construct a large-scale synthetic dataset in this paper.

\section{Proposed Method}
\label{sec:method}

\subsection{Overview}
\label{sec:overview}
The overall pipeline of the proposed method is illustrated in~\cref{fig:method}.
Given an image sequence $x^\mathrm{q}$, whose identity label is denoted as $y^\mathrm{q}$, as query, person Re-ID aims to return a ranking list of gallery image sequences where the sequences from the same person, \textit{i.e.}, $\{x^\mathrm{g}_i|y^\mathrm{g}_i=y^\mathrm{q}\}$, can rank top. 
To this end, this paper proposes a two-branch confidence-aware re-ranking framework.
Specifically, given a query and several galleries, we first extract the appearance and gait features. Base on that, we conduct the inference stage to get the initial ranking list and collect top-K candidates for model training.
Then, we construct candidate relation graph and use GCN to estimate the confidence for balancing the two branches.
Finally, we fuse the confidence-aware re-weighted appearance and gait representations, and re-rank the galleries as the final retrieval result.


\subsection{Two-branch Initial Ranking}
\label{sec:framework}

\noindent \textbf{Two-branch top-K candidates collection.} 
Given a query sequence $x^\mathrm{q}$ and a gallery set $\mathcal{G} =\{x^\mathrm{g}_i|_{i=1}^N\}$, where $N$ represents the number of gallery sequences, the extracted appearance features can be denotes as $\mathbf{a}^\mathrm{q}=\mathrm{f_a}(x^\mathrm{q}|\theta )$ and $\{\mathbf{a}^\mathrm{g}_i=\mathrm{f_a}(x^\mathrm{g}_i|\theta )|_{i=1}^N\}$, respectively.  
Similarly, the gait features can be denoted as $\mathbf{g}^\mathrm{q}=\mathrm{f_g}(x^\mathrm{q}|\phi )$ and $\{\mathbf{g}^\mathrm{g}_i=\mathrm{f_g}(x^\mathrm{g}_i|\phi )|_{i=1}^N\}$. 
$\mathrm{f_a}(\cdot |\theta )$ denotes the appearance feature extraction model with parameters  $\theta $, and $
\mathrm{f_g}(\cdot |\phi )$ denotes the gait model with parameters  $\phi $. 
Then, we can obtain the ranking list, i.e., $\mathcal{R}^\mathrm{a} =\{a_1, a_2, \cdots ,a_N\}$,  through the appearance features according to the pairwise distance $d(\mathbf{a}^\mathrm{q},\mathbf{a}^\mathrm{g}_i)$ between the query $x^\mathrm{q}$ and each gallery $x^\mathrm{g}_i$, where $d(\mathbf{a}^\mathrm{q},\mathbf{a}^\mathrm{g}_{a_i}) < d(\mathbf{a}^\mathrm{q},\mathbf{a}^\mathrm{g}_{a_{i+1}})$. 
In the same way, the ranking list $\mathcal{R}^\mathrm{b} = \{g_1, g_2, \cdots ,g_N\}$ calculated through the gait (biological) features can also be obtained.

Since the value of $N$ is usually large, we select $K$-nearest neighbors of the query for subsequent training.
Specifically, we select the top $\gamma \cdot K \  (0 \leq \gamma \leq 1)$ samples of $\mathcal{R}^\mathrm{a}$, and supplement the rest with the top samples of $\mathcal{R}^\mathrm{b}$. The selected $K$ candidates are denoted as $\mathcal{R}^\mathrm{t}=\{t_1, t_2,\cdots , t_K\}$.
\\
\textbf{Candidate relation graph building.} 
Given the selected samples $\mathcal{R}^\mathrm{t}$, we first construct two graphs with respect to appearance and gait features, respectively.
We approximately represent the appearance similarity between the query $x^\mathrm{q}$ and sample $x^\mathrm{g}_{t_i}$ in $\mathcal{R}^\mathrm{t}$ by applying min-max normalization to the distance:
\begin{equation}
\begin{split}
  s_k^a = \frac{M-d(\mathbf{a}^\mathrm{q},\mathbf{a}^\mathrm{g}_{t_k})} 
  {M-m},
  \label{eq:sim}
\end{split}
\end{equation}
where $M$ and $m$ denote the maximum and minimum values of all $d(\mathbf{a}^\mathrm{q},\mathbf{a}^\mathrm{g}_{t_k})$ with $t_k \in \mathcal{R}^\mathrm{t}$, respectively. The gait similarity $s_k^\mathrm{b}$ can be represented by the same way.

The similarity of appearance features between the query and samples in $\mathcal{R}^\mathrm{t}$ and that of gait features are combined as an initialized similarity matrix $\mathbf{S}_0\in \mathbb{R}^{K\times 2}$, and then embedded into a shared matrix $\mathbf{S}\in \mathbb{R}^{K\times D}$ by an encoder.
For the appearance graph, the nodes represent all samples, and the features of each node is initialized as the vector of the corresponding row in the matrix $\mathbf{S}$. The edges represent the proximity relationship between the samples. If the gait feature $\mathbf{g}^\mathrm{g}_{t_j}$ is one of the $n$-nearest neighbors of feature $\mathbf{g}^\mathrm{g}_{t_k}$, there will be an edge from $t_j$ to $t_k$, and the edge feature $\mathbf{e}_{jk}$ will be defined as the element-wise multiplication of $\mathbf{a}^\mathrm{g}_{t_j}$ and $\mathbf{a}^\mathrm{g}_{t_k}$.
In the same way, the gait graph can be constructed.

\subsection{Confidence Balanced Re-ranking}
\label{sec:conf}
\noindent
\textbf{Confidence estimation.} 
Based on the appearance and gait graph, we aim to estimate confidence for the appearance/gait feature of each sample, which explicitly represents the reliability of the feature.
Inspired by the effectiveness and generalization that GCN has demonstrated in many computer vision applications~\cite{welling2016semi,hamilton2017inductive,shi2019two,wang2022frame,dhingra2021border,xie2022improving} , we adopt GCN here to explore the latent structural relationships among the candidate samples.
The message passing of node $k$ in layer $l$ can be formulated as
\begin{equation}
  \mathbf{H}_k^l = \sigma (\sum _{j \in \mathcal{N}(k) \cup \{k\}} \mathbf{\alpha} ^ \top \mathbf{e}_{jk}\mathbf{W}^{l}\mathbf{H}_k^{l-1}),
  \label{eq:gcn}
\end{equation}
where $\mathbf{H}_k^l$ denotes the feature of node $k$ in layer $l$, $\mathbf{\alpha} ^ \top \mathbf{e}_{jk}$ denotes the weight of the edge from node $j$ to node $k$, and $\mathbf{W}^{l}$ denotes the learnable weight matrix of layer $l$. $\mathcal{N}(k)$ is the set of all nodes that have an edge pointing to node $k$. $\sigma (\cdot)$ is an activation function.

After the GCN layer, a linear layer is used to map the node feature into a one-dimensional confidence score, which can be formulated as
\begin{equation}
  c_k = \omega ^\top \mathbf{H}_k^L,
  \label{eq:conf}
\end{equation}
where $\omega$ denotes the learnable weight of the linear layer, and $c_k$ denotes the estimated confidence score for node $k$.
Note that the network structure used to extract appearance confidences from the appearance graph and that used to extract gait confidences from the gait graph are the same, while do not share parameters.
The appearance confidence for node $k$ can be denoted as $c_k^\mathrm{a}$ and the gait confidence can be denoted as $c_k^\mathrm{b}$.
\\
\textbf{Confidence-aware re-ranking.} 
With the estimated confidence, we propose to adjust the weight of original appearance/gait similarity for each sample in the fusion phase. 
The final similarity between the query $x^\mathrm{q}$ and the sample $x^\mathrm{g}_{t_k}$ in $\mathcal{R}^t$ is defined as:
\begin{equation}
  s_k = c_k^\mathrm{a} s_k^\mathrm{a} +  c_k^\mathrm{b} s_k^\mathrm{b} + \omega _0 ^\top \mathbf{S}(k,:),
  \label{eq:final}
\end{equation}
where $\mathbf{S} (k,:)\in \mathbb{R}^D $ denotes the vector in the \emph{k-th} row of matrix $\mathbf{S}$, \textit{i.e.}, the initialized node feature, which is mapped into one dimension using a linear layer with weight $\omega _0$. 
Finally, the revised ranking list can be obtained by descending sort of the final similarities.
\\
\textbf{Confidence loss.} 
To guide the learning of the network, we propose the confidence loss by designing pseudo-labels for the confidence.
Specifically, assume the ground-truth label of the sample $x^\mathrm{g}_{t_k}$ is $b_k$. When the identity label of sample $x^\mathrm{g}_{t_k}$ equals that of the query, \textit{i.e.}, $y^\mathrm{g}_{t_k}=y^\mathrm{q}$, the value of $b_k$ is 1, otherwise it is 0.
Given the initial appearance similarity $s_k^\mathrm{a}$ and the ground-truth label $b_k$ of sample $x^\mathrm{g}_{t_k}$, we propose to define the pseudo-label of confidence for appearance feature as
\begin{equation}
  \tilde{c}_k^\mathrm{a} = |(1-b_k) - s_k^\mathrm{a}|,
  \label{eq:pseudo}
\end{equation}
where $|\cdot|$ denotes the symbol of absolute value.

The proof is as follows. Since similarity indicates the relational degree between two samples, \textit{i.e.}, the greater the similarity, the higher the probability that the query and the gallery sample have the same identity label, we define a threshold $\lambda $ to represent the prediction results based on similarities. If $s_k^\mathrm{a}\geq \lambda$, we consider the prediction result $\hat{b}_k^\mathrm{a}$ to be 1.
In this case, if $b_k$ is also 1, \textit{i.e.}, the prediction is correct, the value of confidence should be large. Intuitively, the larger the similarity $s_k^\mathrm{a}$, the larger the confidence $\tilde{c}_k^\mathrm{a}$ should be. Therefore, $\tilde{c}_k^\mathrm{a}$ can be assigned the value of $s_k^\mathrm{a}$.
In contrast, if $b_k=0$, \textit{i.e.}, the prediction is wrong, the value of confidence should be small. And the larger the similarity $s_k^\mathrm{a}$, the smaller the confidence $\tilde{c}_k^\mathrm{a}$ should be. Therefore, $\tilde{c}_k^\mathrm{a}$ can be assigned the value of $1-s_k^\mathrm{a}$.
The above can be formulated as
\begin{equation}
  \tilde{c}_k^\mathrm{a} = \begin{cases}
  s_k^\mathrm{a}, & b_k=1 \\
  1 -s_k^\mathrm{a}, & b_k=0 \\
  \end{cases} .
  \label{eq:proof}
\end{equation}
Similarly, if $s_k^\mathrm{a}< \lambda$, we consider the prediction result $\hat{b}_k^\mathrm{a} $ to be 0. In this case, if $b_k=1$, \textit{i.e.}, the prediction is wrong, the value of confidence should be small. And the smaller the similarity $s_k^\mathrm{a}$, the smaller the confidence $\tilde{c}_k^\mathrm{a}$ should be. Therefore, $\tilde{c}_k^\mathrm{a}$ can be assigned the value of $s_k^\mathrm{a}$.
In contrast, if $b_k=0$, \textit{i.e.}, the prediction is correct, the value of confidence should be large. And the smaller the similarity $s_k^\mathrm{a}$, the larger the confidence $\tilde{c}_k^\mathrm{a}$ should be. Therefore, $\tilde{c}_k^\mathrm{a}$ can be assigned the value of $1-s_k^\mathrm{a}$.
The formulation of the above analysis takes the same form as~\cref{eq:proof}.
Further, \cref{eq:proof} can be rewritten as follows:
\begin{equation}
  \tilde{c}_k^\mathrm{a} = \begin{cases}
  s_k^\mathrm{a} - (1-b_k), & b_k=1 \\
  (1-b_k) -s_k^\mathrm{a}, & b_k=0 \\
  \end{cases} ,
  \label{eq:proof2}
\end{equation}
which is exactly the form of classified discussion of~\cref{eq:pseudo}.
In the same way, the pseudo-label of confidence for gait feature can be defined as
\begin{equation}
  \tilde{c}_k^\mathrm{b} = |(1-b_k) - s_k^\mathrm{b}|.
  \label{eq:g-pseudo}
\end{equation}

In the training stage, the designed pseudo-labels $\tilde{c}_k^\mathrm{a}$ and $\tilde{c}_k^\mathrm{b}$ play a role of ground-truths of confidence to guide the confidence estimation.
The confidence loss can be formulated as
\begin{equation}
  \mathcal{L}_c = \sum _{k=1}^K (|\tilde{c}_k^\mathrm{a} - c_k^\mathrm{a}| + |\tilde{c}_k^\mathrm{b} - c_k^\mathrm{b}|).
  \label{eq:confloss}
\end{equation}

\subsection{Network Setting}
\label{sec:imple}
\noindent
\textbf{Ranking loss.} 
We also consider the widely used triplet ranking loss~\cite{hermans2017defense} in person Re-ID: 
\begin{equation}
  \mathcal{L}_r = \sum _{k=1}^{K_p}  \sum _{j=1}^{K_n} [\epsilon-(s_k^\mathrm{p}-s_j^\mathrm{n})]_+,
  \label{eq:rankloss}
\end{equation}
where $s_k^\mathrm{p}$ denotes the final similarity between the query and the positive sample whose ground-truth label is 1, and $s_j^\mathrm{n}$ denotes that for negative sample whose ground-truth label is 0. $K_p$ and $K_n$ denote the number of positive and negative samples, respectively, with $K_p+K_n=K$. $\epsilon$ denotes the margin.
Combining the confidence and ranking loss, the total loss in the training stage can be formulated as $\mathcal{L} = \mathcal{L}_c+\mathcal{L}_r$.
\\
\textbf{Implementation Details.}
We use SINet~\cite{bai2022salient} proposed for video-based person Re-ID as the appearance model $\mathrm{f_a}(\cdot |\theta )$, and GaitSet~\cite{chao2021gaitset} proposed for gait recognition as the gait model $\mathrm{f_g}(\cdot |\phi )$. 
For SINet, we directly input RGB sequences for appearance feature extraction. While for GaitSet, we first perform an instance segmentation algorithm HTC~\cite{chen2019hybrid} to extract silhouettes from RGB sequences, and then input silhouette sequences for gait feature extraction.
SINet and GaitSet are pre-trained on ImageNet~\cite{deng2009imagenet} and GREW~\cite{zhu2021gait}, respectively, and then trained on the corresponding CCVReID datasets to get the initial features.
The number of selected samples for training, i.e., $K$, is set to 100.
The ratio $\gamma$ of selecting samples from $\mathcal{R}^\mathrm{a}$ is set to 0.75.
In graph building, following~\cite{xie2022improving}, the encoder takes a Multi-Layer Perceptron (MLP) structure, which consists of a linear, BN, PReLU, dropout, and linear layer successively. The dimension of the initialized node feature $D$ is 32. The $n$ in $n$-nearest neighbors for edge connection is set to 30. In confidence estimation, one GCN layer is used, with the output dimension of 32. The margin $\epsilon$ in~\cref{eq:rankloss} is set to 0.2. We use Adam~\cite{kingma2015adam} algorithm to optimize the model in all experiments.
During the testing phase, given a query, the $K$ samples selected for training from gallery are ranked in the top-K according to their final similarities, followed by the rest samples reordered according to the summation of the initial appearance rank in $\mathcal{R}^\mathrm{a}$ and gait rank in $\mathcal{R}^\mathrm{b}$.

\begin{figure}
  \centering
  \begin{subfigure}{0.32\linewidth}
    \includegraphics[width=0.9\linewidth]{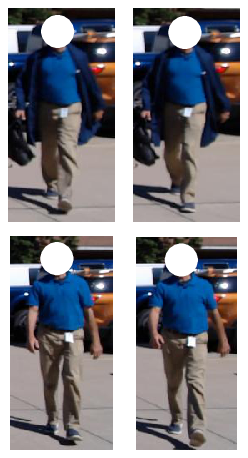}
    \caption{CCVID.}
    \label{fig:dataset-a}
  \end{subfigure}
  \begin{subfigure}{0.32\linewidth}
    \includegraphics[width=0.9\linewidth]{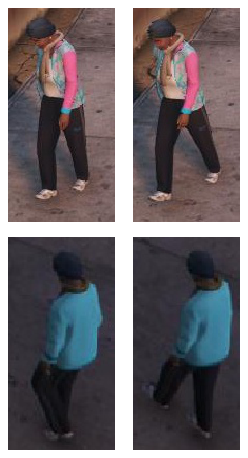}
    \caption{SCCVReID.}
    \label{fig:dataset-b}
  \end{subfigure}
  \begin{subfigure}{0.3\linewidth}
    \includegraphics[width=0.9\linewidth]{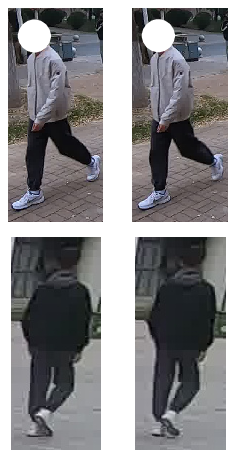}
    \caption{RCCVReID.}
    \label{fig:dataset-c}
  \end{subfigure}
  \caption{Examples of CCVID~\cite{gu2022clothes} and the proposed benchmark datasets in this paper.}
  \label{fig:dataset}
  \vspace{-10pt}
\end{figure}

\begin{table*}[]
\centering
\begin{tabular}{cccccccccc}
\hline
\multirow{2}{*}{} & \multirow{2}{*}{Methods} & \multicolumn{4}{c}{CC} & \multicolumn{4}{c}{Standard} \\  \cline{3-10}
 &  & mAP & R1 & R5 & R10 & mAP & R1 & R5 & R10 \\ \hline
\multirow{3}{*}{\begin{tabular}[c]{@{}c@{}}Video-based w/o\\ clothes-changing\end{tabular}} & AP3D~\cite{gu2020appearance} & 26.7 & 47.1 & 71.5 & 80.2 & 39.1 & 79.4 & 94.3 & 96.9 \\ 
 & TCLNet~\cite{hou2020temporal} & 30.1 & 48.6 & 72.4 & 81.7 & 43.2 & 84.3 & 95.6 & 97.3 \\
 & SINet~\cite{bai2022salient} & 33.0 & 51.0 & 75.5 & 84.6 & 46.5 & 84.4 & \textbf{97.4} & \textbf{98.6} \\ \hline
\multirow{3}{*}{Gait recognition} & GaitSet~\cite{chao2021gaitset} & 16.7 & 23.4 & 56.0 & 68.5 & 20.6 & 33.6 & 68.4 & 80.1 \\
 & GaitPart~\cite{fan2020gaitpart} & 15.9 & 22.4 & 54.7 & 68.8 & 19.2 & 30.4 & 64.7 & 79.3 \\ 
 & GaitGL~\cite{lin2021gait} & 11.7 & 15.6 & 44.3 & 56.7 & 14.5 & 23.5 & 54.1 & 65.4 \\ \hline
\multirow{6}{*}{\begin{tabular}[c]{@{}c@{}}Image-based\\ clothes-changing\end{tabular}} & ReIDCaps-R~\cite{huang2019beyond} & 12.2 & 30.0 & 53.5 & 65.1 & 12.9 & 32.2 & 56.0 & 67.0 \\
 & ReIDCaps-A~\cite{huang2019beyond} & 10.2 & 26.0 & 48.6 & 60.4 & 21.2 & 66.8 & 87.1 & 92.1 \\
 & Pixel\_sampling-R~\cite{shu2021semantic} & 13.8 & 29.8 & 57.3 & 68.9 & 21.7 & 57.9 & 81.8 & 88.3 \\
 & Pixel\_sampling-A~\cite{shu2021semantic} & 10.1 & 23.2 & 46.8 & 60.4 & 17.4 & 50.4 & 77.2 & 85.8 \\ 
 & GI-ReID-R~\cite{jin2022cloth} & 4.9 & 8.9 & 21.4 & 31.3 & 6.4 & 12.4 & 30.1 & 42.2 \\
 & GI-ReID-A~\cite{jin2022cloth} & 8.2 & 14.3 & 33.3 & 43.1 & 8.5 & 14.7 & 35.1 & 44.5 \\ \hline
\multirow{2}{*}{\begin{tabular}[c]{@{}c@{}}Video-based\\ clothes-changing\end{tabular}} & CAL~\cite{gu2022clothes} & 31.2 & 48.9 & 71.3 & 81.8 & 45.4 & \textbf{87.6} & 95.0 & 96.9 \\
 & Ours & \textbf{39.8} & \textbf{54.8} & \textbf{76.2} & \textbf{86.1} & \textbf{51.1} & 83.0 & 94.5 & 97.4 \\ \hline
\end{tabular} 
\caption{Comparison with the state-of-the-art Re-ID and gait recognition methods on SCCVReID dataset (\%).}
\label{tab:compar}
\vspace{-10pt}
\end{table*}

\section{Datasets}
\label{sec:data}
Current works focus little on CCVReID. Of the datasets available for this task, including Motion-ReID~\cite{zhang2018long}, CVID-reID~\cite{zhang2020learning}, and CCVID~\cite{gu2022clothes}, only CCVID is publicly available.
However, as seen in~\cref{fig:dataset-a}, all samples in the CCVID were taken from the same view, \textit{i.e.}, the frontal view, and have clean backgrounds and no occlusion, which is not fully applicable to real-world scenes.
Therefore, to advance related researches, we build two new benchmark datasets in this paper, including a large-scale synthetic one named SCCVReID, and a small real one named RCCVReID. 
We will give a brief introduction in~\cref{sec:synthetic,sec:real}.

\subsection{The SCCVReID Dataset}
\label{sec:synthetic}

Following~\cite{fabbri2018learning}, we collect a large-scale synthetic dataset for CCVReID in surveillance scenarios by exploiting the highly photorealistic video game \emph{Grand Theft Auto V}.
The examples can be seen in~\cref{fig:dataset-b}.
We set up 10 surveillance cameras within 5 scenes (2 cameras for each scene) to collect data. 
After recording the pedestrians for 48 hours in the game at 60 FPS, we use the automatic bounding boxes to crop out RGB sequences of each person.
The obtained SCCVReID dataset contains 9,620 sequences from 333 identities and each identity has 2-37 suits of clothes, with an average of 7.
Each sequence contains a number of frames ranging between 8 and 165 with an average length of 37.
For evaluation, 167 identities with 5,768 sequences are used for training and the rest 166 identities with 3,852 sequences for testing. In the test set, we select the first sequence of each suit of each identity as query.
In total, 971 sequences are used as query, and the rest 2,881 as gallery.

\subsection{The RCCVReID Dataset}
\label{sec:real}
We also collect a real dataset named RCCVReID to support researches on CCVReID, which can be seen in~\cref{fig:dataset-c}. 
All raw videos were recorded in outdoor scenarios. For each video, we first performed ByteTrack~\cite{zhang2022bytetrack} to generate human bounding boxes with unified IDs, and used them to crop out RGB sequences of each identity, which is then split into several short sequences (about 200 frames).
After that, we merged the sequences of the same identity from different videos and manually labeled the clothes IDs. 
In total, 6,948 sequences from 34 identities are obtained.
Each identity has 2-9 suits of clothes, with an average of 4.
We consider two evaluation settings on this dataset.
(1) Due to the small number of identities in RCCVReID, we train the models on the large-scale synthetic dataset SCCVReID and only use RCCVReID for performance evaluation. In this case, 2,133 sequences are used as query, and the rest 4,815 sequences as gallery.
(2) Despite the small number of identities, the total sample size is sufficient for training models.
Thus, we divide the training and test sets as existing Re-ID datasets do.
Specifically, 20 identities with 5,532 sequences are reserved for training, and the remaining 14 identities are used for test.
In the test set, 486 sequences are used as query, and the rest 932 sequences as gallery.

\section{Experiments}
\label{sec:exp}

\subsection{Datasets and Evaluation Protocol}
\label{sec:eval}

We perform experiments on three CCVReID datasets, \textit{i.e.}, CCVID~\cite{gu2022clothes}, SCCVReID, and RCCVReID, to evaluate the proposed method. 
For evaluation, we focus on two kinds of test settings, \textit{i.e.}, clothes-changing (CC) setting and standard setting.
In clothes-changing setting, gallery samples that have the same identity label and clothes label with query are removed, that is, only gallery samples with clothes-changing are considered.
While in standard setting, both clothes-consistent and clothes-changing samples are used as gallery to calculate accuracy.
Following existing person Re-ID works, we adopt the Rank-1/5/10 (R1/R5/R10) of CMC curve and the mAP (mean average precision) score as the evaluation protocols.

\begin{table*}[]
\centering
\begin{tabular}{cccccccccc}
\hline
\multirow{3}{*}{} & \multirow{3}{*}{Methods} & \multicolumn{4}{c}{Only test} & \multicolumn{4}{c}{With train} \\ \cline{3-10} 
 &  & \multicolumn{2}{c}{CC} & \multicolumn{2}{c}{Standard} & \multicolumn{2}{c}{CC} & \multicolumn{2}{c}{Standard} \\ \cline{3-10} 
 &  & mAP & R1 & mAP & R1 & mAP & R1 & mAP & R1 \\ \hline
\multirow{3}{*}{\begin{tabular}[c]{@{}c@{}}Video-based w/o\\ clothes-changing\end{tabular}} & AP3D~\cite{gu2020appearance} & 9.2 & 16.6 & 30.0 & 85.3 & 21.0 & 20.2 & 62.7 & 94.9 \\
 & TCLNet~\cite{hou2020temporal} & 12.0 & 20.2 & 37.3 & 95.6 & 22.6 & 28.0 & 62.3 & 95.7 \\
 & SINet~\cite{bai2022salient} & 8.2  & 15.2 & 29.8 & 86.9 & 23.8 & 31.9 & 67.3 & 96.7 \\ \hline
\multirow{3}{*}{Gait recognition} & GaitSet~\cite{chao2021gaitset} & 6.3  & 13.0 & 16.5 & 66.5 & 16.4 & 26.5 & 43.2 & 84.4 \\
 & GaitPart~\cite{fan2020gaitpart} & 6.5  & 14.6 & 16.3 & 64.2 & 15.6 & 23.7 & 37.4 & 77.4 \\
 & GaitGL~\cite{lin2021gait} & 5.0  & 10.6 & 10.1 & 44.6 & 18.7 & 31.1 & 45.3 & 87.9 \\ \hline
\multirow{6}{*}{\begin{tabular}[c]{@{}c@{}}Image-based\\ clothes-changing\end{tabular}} & ReIDCaps-R~\cite{huang2019beyond} & 7.3  & 13.2 & 25.5 & 85.0 & 18.0 & 22.0 & 56.1 & 95.5 \\
 & ReIDCaps-A~\cite{huang2019beyond} & 8.2  & 14.3 & 29.8 & 90.8 & 21.6 & 29.6 & 62.4 & 97.3 \\
 & Pixel\_sampling-R~\cite{shu2021semantic} & 4.5  & 4.5  & 9.1  & 14.6 & 21.3 & 27.2 & 55.3 & 87.4 \\
 & Pixel\_sampling-A~\cite{shu2021semantic} & 6.0  & 11.7 & 20.7 & 74.5 & 17.9 & 17.7 & 58.0 & 92.2 \\
 & GI-ReID-R~\cite{jin2022cloth} & 4.5  & 3.9  & 9.9  & 18.7 & 11.5 & 9.1  & 31.8 & 44.0 \\
 & GI-ReID-A~\cite{jin2022cloth} & 5.2  & 5.7  & 11.9 & 21.8 & 14.1 & 11.5 & 37.6 & 53.5 \\ \hline
\multirow{2}{*}{\begin{tabular}[c]{@{}c@{}}Video-based\\ clothes-changing\end{tabular}} & CAL~\cite{gu2022clothes} & 13.0 & 20.9 & 40.3 & \textbf{96.3} & 26.0 & 31.5 & 68.1 & 99.2 \\
 & Ours & \textbf{14.5} & \textbf{25.7} & \textbf{43.1} & 95.8 & \textbf{33.9} & \textbf{46.7} & \textbf{72.3} & \textbf{99.4} \\ \hline
\end{tabular}
\caption{Comparison with the state-of-the-art Re-ID and gait recognition methods on RCCVReID dataset (\%).}
\label{tab:rccvreid}
\vspace{-10pt}
\end{table*}

\begin{table}[]
\centering
\resizebox{0.98\linewidth}{!}{
\begin{tabular}{cccccccccc}
\hline
\multirow{2}{*}{} & \multirow{2}{*}{Methods} & \multicolumn{2}{c}{CC} & \multicolumn{2}{c}{Standard} \\ \cline{3-6} 
 &  & mAP & R1 & mAP & R1 \\ \hline
\multirow{3}{*}{VReID} & AP3D~\cite{gu2020appearance} & 69.4 & 75.5 & 71.7 & 77.0\\
 & TCLNet~\cite{hou2020temporal} & 71.2 & 77.1 & 73.3 & 78.3 \\
 & SINet~\cite{bai2022salient} & 77.8 & 81.1 & 82.4 & 85.6 \\ \hline
\multirow{3}{*}{GR} & GaitSet~\cite{chao2021gaitset} & 62.6 & 72.7 & 69.2 & 77.6 \\
 & GaitPart~\cite{fan2020gaitpart} & 60.1 & 73.1 & 67.0 & 79.9 \\
 & GaitGL~\cite{lin2021gait} & 69.0 & 82.1 & 75.4 & 88.1 \\ \hline
\multirow{6}{*}{CCIReID} & ReIDCaps-R~\cite{huang2019beyond} & 45.9 & 54.2 & 49.8 & 58.6 \\
 & ReIDCaps-A~\cite{huang2019beyond} & 46.2 & 50.1 & 49.9 & 52.5  \\
 & Pixel\_sampling-R~\cite{shu2021semantic} & 42.4 & 52.5 & 48.0 & 58.2 \\
 & Pixel\_sampling-A~\cite{shu2021semantic} & 50.4 & 57.3 & 55.7 & 62.9  \\
 & GI-ReID-R~\cite{jin2022cloth} & 17.6 & 17.3 & 21.3 & 23.1  \\
 & GI-ReID-A~\cite{jin2022cloth} & 27.3 & 27.8 & 31.7 & 34.7 \\ \hline
\multirow{2}{*}{CCVReID} & CAL~\cite{gu2022clothes} & 81.7 & 83.8 & 83.2 & 84.5 \\
 & Ours & \textbf{84.5} & \textbf{88.1} & \textbf{87.1} & \textbf{89.7} \\ \hline
\end{tabular}}
\caption{Comparison with the state-of-the-art Re-ID and gait recognition methods on CCVID~\cite{gu2022clothes} dataset (\%).}
\label{tab:ccvid}
\vspace{-10pt}
\end{table}

\subsection{Comparison with State-of-the-art Methods}
\label{sec:comp}
Since there are few works to explore CCVReID task, we compare our method with four kinds of methods for a comprehensive evaluation: 1) video-based person Re-ID methods that do not involve clothes-changing, including AP3D~\cite{gu2020appearance}, TCLNet~\cite{hou2020temporal}, and SINet~\cite{bai2022salient}, 2) gait recognition methods, including GaitSet~\cite{chao2021gaitset}, GaitPart~\cite{fan2020gaitpart}, and GaitGL~\cite{lin2021gait}, 3) image-based clothes-changing methods, including ReIDCaps~\cite{huang2019beyond}, Pixel\_sampling~\cite{shu2021semantic}, and GI-ReID~\cite{jin2022cloth}, 4) video-based clothes-changing method CAL~\cite{gu2022clothes}.
Note that image-based methods receive a single frame as input, we report the results under two different settings of such methods, \textit{i.e.}, randomly select a frame for feature extraction (denoted with suffix "-R"), and extract the features of all frames and take the average as the final feature (denoted with suffix "-A").
\\
\textbf{Results on SCCVReID.}
As shown in~\cref{tab:compar}, compared with image-based methods, video-based Re-ID and gait recognition methods achieve better performance, which indicates the importance of spatio-temporal information in improving the accuracy of Re-ID.
Besides, our method achieves the best performance under CC setting.
Compared with CAL~\cite{gu2022clothes}, our method improves the mAP and R1 by more than 8.6\% and 5.9\%, respectively.
However, it is observed that under standard setting, our method is lower than SINet~\cite{bai2022salient} in R1, R5, and R10. 
Intuitively, this can be attributed to the low accuracy of the gait features.
While gait information can assist in identification when appearance information is unreliable, \textit{i.e.}, with clothes-changing, it may also introduce certain interference when appearance information are reliable, \textit{i.e.}, with clothes-consistent.
Despite this, our method achieve the best mAP under this setting, that is, our method allows the highest average ranking for all gallery samples with the same identity as the query.
\\
\textbf{Results on RCCVReID.}
We compare our method with state-of-the-art methods under two evaluation settings, \textit{i.e.}, use SCCVReID for training and use RCCVReID only for testing, or use RCCVReID for both training and testing.
The results are shown in~\cref{tab:rccvreid}.
Note that the only test setting can be considered as cross-domain CCVReID, evaluating the generalization ability of the models.
Under both evaluation settings, our method achieves the best mAP and R1 under CC setting, and the best mAP under standard setting, which are consistent with the results on SCCVReID.
\\
\textbf{Results on CCVID.}
As seen in~\cref{tab:ccvid}, our method outperforms all the other methods by a large margin on all evaluation protocols.
Since CCVID is built based on gait recognition dataset, the extracted gait features are relatively more discriminative.
Thus, the drop in R1 under standard setting caused by the interference of low-quality gait features does not exist here.

\subsection{Ablation Study}
\label{sec:abl}
In this section, we conduct a series of ablation experiments on SCCVReID to verify each design of our method.
\\
\textbf{The effectiveness of proposed method.}
To verify the effectiveness of the proposed method, we compare it with different fusion framework in~\cref{tab:abl}.
The top 2 rows report the performance of using appearance and gait features solely, respectively.
The third row report the performance when the initial appearance and gait similarity are directly summed for re-ranking.
SEF~\cite{xie2022improving} is the inspiration for this paper. Differently, SEF considers the appearance (gait) features only when building appearance (gait) graph, and uses GCN to directly predict the final similarity.
The results show that all fusion strategies bring an improvement in accuracy under CC setting and mAP under standard setting, while a certain drop in R1 under standard setting.
The main reason for that has been discussed in~\cref{sec:comp}.
However, compared to the other two fusion strategies, our method obtains the highest improvement under clothes-changing setting and the least drop of R1 under standard setting, which indicates the superiority of our method.
\begin{table}[] 
\centering
\resizebox{0.98\linewidth}{!}{
\begin{tabular}{lcccc}
\hline
 & \multicolumn{2}{c}{CC} & \multicolumn{2}{c}{Standard} \\ \cline{2-5} 
 & mAP & R1 & mAP & R1 \\ \hline
App only & 33.0 & 51.0 & 46.5 & 84.4 \\
Gait only & 18.7 & 28.1 & 23.3 & 38.0 \\
Direct summation & 36.4 & 50.2 & 48.7 & 81.7 \\
SEF~\cite{xie2022improving} & 38.8 & 50.6 & 47.2 & 67.1 \\
Ours w pseudo-label (Oracle) & 69.4 & 97.2 & 75.2 & 99.6 \\
Ours w/o appConf & 39.4 & 53.6 & 50.7 & 81.7 \\
Ours w/o gaitConf & 39.2 & 52.6 & 50.8 & 82.3 \\
Ours & 39.8 & 54.8 & 51.1 & 83.0 \\
\hline
\end{tabular}}
\caption{Analysis of different fusion framework and the effectiveness of the proposed confidence loss (\%).}
\label{tab:abl}
\vspace{-10pt}
\end{table}
\\
\textbf{The effectiveness of confidence loss.}
To evaluate the effectiveness of the proposed confidence loss, We first list the accuracy obtained by calculating the final similarity using designed pseudo-label of confidence.
As shown in~\cref{tab:abl}, compared with the simple summation of similarities, our model with pseudo-label, \textit{i.e.}, the ground-truth of confidence, improves 33.0\% on mAP and 47.0\% on R1 under CC setting, 26.5\% on mAP and 17.9\% on R1 under standard setting. 
The results demonstrate the validity of the designed pseudo-label of confidence, which can be regarded as the oracle of the proposed confidence-aware method.
We also compare our method with and without appearance/gait confidence loss. In~\cref{tab:abl}, we can observe that removing the confidence loss of both the appearance and gait branch causes a certain degradation under both clothes-changing and standard setting. 
This demonstrates that the proposed loss can effectively supervise the learning of confidence.
\\
\textbf{Discussion of two-branch framework.}
We conduct ablation experiments to verify each branch of the proposed framework in~\cref{tab:branch}.
We discuss three settings of confidence: 1) the confidence are output by the network, which is represented by $\surd$. 2) the confidence is fixed to 0, that is, the original similarity is not used to calculate the final similarity, only the other branch is considered. However, in this case, the original similarity is not completely neglected, since the information has been implicitly included in the linear layer.
3) the confidence is fixed to 1, that is, the original similarity is used to calculate the final similarity without weight adjustment.
From the comparison between the first and second rows, the third and fourth rows, we can see that the model with confidence set to 1 outperforms the model with confidence set to 0. That shows that the importance of the original similarity in the re-ranking stage.
From the comparison between the top 4 rows and the fifth to the eighth rows, we can see that the model with confidence fixed for one branch outperforms the model with confidence fixed for both branches. Also, the model without confidence fixed achieves the best performance (the last row), which fully proves the superiority of the proposed network for confidence learning.
Besides, we compare the models with and without linear layer (the last 2 rows).
The results show that with addition of the linear layer, the performance of model is improved.
\begin{table}[]
\centering
\resizebox{0.98\linewidth}{!}{
\begin{tabular}{ccccccc}
\hline
\multirow{2}{*}{AppConf} & \multirow{2}{*}{GaitConf} & \multirow{2}{*}{Linear} & \multicolumn{2}{c}{CC} & \multicolumn{2}{c}{Standard} \\ \cline{4-7} 
& & & mAP & R1 & mAP & R1 \\ \hline
0 & $\surd$ & $\surd$ & 38.4 & 51.7 & 50.5 & 82.7 \\
1 & $\surd$ & $\surd$ & 38.7 & 52.3 & 50.9 & 83.8 \\
$\surd$ & 0 & $\surd$ & 37.6 & 52.1 & 49.8 & 83.0 \\
$\surd$ & 1 & $\surd$ & 37.6 & 52.2 & 50.0 & \textbf{84.3} \\
0 & 0 & $\surd$ & 36.5 & 50.9 & 49.1 & 82.3 \\
0 & 1 & $\surd$ & 36.3 & 49.2 & 48.3 & 76.7 \\
1 & 0 & $\surd$ & 36.5 & 50.7 & 49.0 & 81.1 \\
1 & 1 & $\surd$ & 36.6 & 50.8 & 49.2 & 82.1 \\
$\surd$ & $\surd$ &  & 38.9 & 52.7 & 50.7 & 83.2 \\
$\surd$ & $\surd$ & $\surd$ & \textbf{39.8} & \textbf{54.8} & \textbf{51.1} & 83.0 \\ \hline
\end{tabular}}
\caption{Results of different settings in the two-branch framework (\%). ``0'' means the confidence is set to 0, \textit{i.e.}, the corresponding branch is removed. ``1'' means the confidence is set to 1, \textit{i.e.}, the original corresponding branch is directly added without weight adjustment. ``$\surd$'' means the confidence is output by the proposed network.}
\label{tab:branch}
\vspace{-10pt}
\end{table}
\begin{table}[]
\centering
\begin{tabular}{lcccc}
\hline
 & \multicolumn{2}{c}{CC} & \multicolumn{2}{c}{Standard} \\ \cline{2-5} 
 & mAP & R1 & mAP & R1 \\ \hline
$\gamma =0$ & 32.4 & 50.3 & 41.9 & 77.2 \\
$\gamma =0.25$ & 36.6 & 53.1 & 48.0 & 81.2 \\
$\gamma =0.5$ & 38.0 & 52.7 & 49.5 & 82.9 \\
$\gamma =0.75$ (Ours) & \textbf{39.8} & \textbf{54.8} & \textbf{51.1} & \textbf{83.0}  \\
$\gamma =1$ & 39.2 & 53.2 & 50.9 & 82.4 \\ \hline
\end{tabular}
\caption{Results of different candidates collection strategy (\%).}
\label{tab:gamma}
\vspace{-10pt}
\end{table}
\\
\textbf{Discussion of candidates collection.}
The parameter $\gamma$ controls the contribution of initial appearance and gait ranking list when selecting training samples.
As shown in~\cref{tab:gamma}, we observe that as the value of $\gamma$ increases, the accuracy under both the settings show a trend of first increasing and then decreasing.
The model achieves the best performance when $\gamma =0.75$.
This means that the optimal neighbors collection strategy is to select most of the samples based on more reliable appearance features, while samples with similar gait features to the query are also considered.


\section{Conclusion}
\label{sec:con}
In this paper, we have focused on a less studied yet and practical problem of video-based person re-identification with clothes-changing (CCVReID).
For this problem, we propose a two-branch confidence-aware re-ranking framework, which fuse the appearance and gait features that have been re-weighted by confidence, for final re-ranking.
We also design confidence pseudo-labels to supervise confidence learning.
Besides, we build two new benchmarks for CCVReID problem, including a large-scale synthetic one and a real-world one.
With the above efforts, we hope to promote the researches on this practical topic.

%

%

{\small
\bibliographystyle{ieee_fullname}
\bibliography{egbib}

\begin{thebibliography}{10}\itemsep=-1pt

\bibitem{bai2022salient}
Shutao Bai, Bingpeng Ma, Hong Chang, Rui Huang, and Xilin Chen.
\newblock Salient-to-broad transition for video person re-identification.
\newblock In {\em CVPR}, pages 7339--7348, 2022.

\bibitem{bak2018domain}
Slawomir Bak, Peter Carr, and Jean-Francois Lalonde.
\newblock Domain adaptation through synthesis for unsupervised person
  re-identification.
\newblock In {\em ECCV}, pages 189--205, 2018.

\bibitem{chao2021gaitset}
Hanqing Chao, Kun Wang, Yiwei He, Junping Zhang, and Jianfeng Feng.
\newblock Gaitset: Cross-view gait recognition through utilizing gait as a deep
  set.
\newblock 44(7):3467--3478, 2021.

\bibitem{chen2022keypoint}
Di Chen, Andreas Doering, Shanshan Zhang, Jian Yang, Juergen Gall, and Bernt
  Schiele.
\newblock Keypoint message passing for video-based person re-identification.
\newblock In {\em AAAI}, volume~36, pages 239--247, 2022.

\bibitem{chen2018video}
Dapeng Chen, Hongsheng Li, Tong Xiao, Shuai Yi, and Xiaogang Wang.
\newblock Video person re-identification with competitive snippet-similarity
  aggregation and co-attentive snippet embedding.
\newblock In {\em CVPR}, pages 1169--1178, 2018.

\bibitem{chen2021learning}
Jiaxing Chen, Xinyang Jiang, Fudong Wang, Jun Zhang, Feng Zheng, Xing Sun, and
  Wei-Shi Zheng.
\newblock Learning 3d shape feature for texture-insensitive person
  re-identification.
\newblock In {\em CVPR}, pages 8146--8155, 2021.

\bibitem{chen2019hybrid}
Kai Chen, Jiangmiao Pang, Jiaqi Wang, Yu Xiong, Xiaoxiao Li, Shuyang Sun,
  Wansen Feng, Ziwei Liu, Jianping Shi, Wanli Ouyang, et~al.
\newblock Hybrid task cascade for instance segmentation.
\newblock In {\em CVPR}, pages 4974--4983, 2019.

\bibitem{chung2017two}
Dahjung Chung, Khalid Tahboub, and Edward~J Delp.
\newblock A two stream siamese convolutional neural network for person
  re-identification.
\newblock In {\em CVPR}, pages 1983--1991, 2017.

\bibitem{deng2009imagenet}
Jia Deng, Wei Dong, Richard Socher, Li-Jia Li, Kai Li, and Li Fei-Fei.
\newblock Imagenet: A large-scale hierarchical image database.
\newblock In {\em CVPR}, pages 248--255. Ieee, 2009.

\bibitem{dhingra2021border}
Naina Dhingra, George Chogovadze, and Andreas Kunz.
\newblock Border-seggcn: improving semantic segmentation by refining the border
  outline using graph convolutional network.
\newblock In {\em ICCV}, pages 865--875, 2021.

\bibitem{eom2021video}
Chanho Eom, Geon Lee, Junghyup Lee, and Bumsub Ham.
\newblock Video-based person re-identification with spatial and temporal memory
  networks.
\newblock In {\em ICCV}, pages 12036--12045, 2021.

\bibitem{fabbri2018learning}
Matteo Fabbri, Fabio Lanzi, Simone Calderara, Andrea Palazzi, Roberto Vezzani,
  and Rita Cucchiara.
\newblock Learning to detect and track visible and occluded body joints in a
  virtual world.
\newblock In {\em ECCV}, pages 430--446, 2018.

\bibitem{fan2020gaitpart}
Chao Fan, Yunjie Peng, Chunshui Cao, Xu Liu, Saihui Hou, Jiannan Chi, Yongzhen
  Huang, Qing Li, and Zhiqiang He.
\newblock Gaitpart: Temporal part-based model for gait recognition.
\newblock In {\em CVPR}, pages 14225--14233, 2020.

\bibitem{fan2020learning}
Lijie Fan, Tianhong Li, Rongyao Fang, Rumen Hristov, Yuan Yuan, and Dina
  Katabi.
\newblock Learning longterm representations for person re-identification using
  radio signals.
\newblock In {\em CVPR}, pages 10699--10709, 2020.

\bibitem{fu2019sta}
Yang Fu, Xiaoyang Wang, Yunchao Wei, and Thomas Huang.
\newblock Sta: Spatial-temporal attention for large-scale video-based person
  re-identification.
\newblock In {\em AAAI}, volume~33, pages 8287--8294, 2019.

\bibitem{gu2022clothes}
Xinqian Gu, Hong Chang, Bingpeng Ma, Shutao Bai, Shiguang Shan, and Xilin Chen.
\newblock Clothes-changing person re-identification with rgb modality only.
\newblock In {\em CVPR}, pages 1060--1069, 2022.

\bibitem{gu2020appearance}
Xinqian Gu, Hong Chang, Bingpeng Ma, Hongkai Zhang, and Xilin Chen.
\newblock Appearance-preserving 3d convolution for video-based person
  re-identification.
\newblock In {\em ECCV}, pages 228--243. Springer, 2020.

\bibitem{hamilton2017inductive}
Will Hamilton, Zhitao Ying, and Jure Leskovec.
\newblock Inductive representation learning on large graphs.
\newblock In {\em NeurIPS}, 2017.

\bibitem{hermans2017defense}
Alexander Hermans, Lucas Beyer, and Bastian Leibe.
\newblock In defense of the triplet loss for person re-identification.
\newblock {\em arXiv preprint arXiv:1703.07737}, 2017.

\bibitem{hoffmann2019learning}
David~T Hoffmann, Dimitrios Tzionas, Michael~J Black, and Siyu Tang.
\newblock Learning to train with synthetic humans.
\newblock In {\em GCPR}, pages 609--623. Springer, 2019.

\bibitem{hong2021fine}
Peixian Hong, Tao Wu, Ancong Wu, Xintong Han, and Wei-Shi Zheng.
\newblock Fine-grained shape-appearance mutual learning for cloth-changing
  person re-identification.
\newblock In {\em CVPR}, pages 10513--10522, 2021.

\bibitem{hou2020temporal}
Ruibing Hou, Hong Chang, Bingpeng Ma, Shiguang Shan, and Xilin Chen.
\newblock Temporal complementary learning for video person re-identification.
\newblock In {\em ECCV}, pages 388--405. Springer, 2020.

\bibitem{huang2019beyond}
Yan Huang, Jingsong Xu, Qiang Wu, Yi Zhong, Peng Zhang, and Zhaoxiang Zhang.
\newblock Beyond scalar neuron: Adopting vector-neuron capsules for long-term
  person re-identification.
\newblock {\em IEEE TCSVT}, 30(10):3459--3471, 2019.

\bibitem{jin2022cloth}
Xin Jin, Tianyu He, Kecheng Zheng, Zhiheng Yin, Xu Shen, Zhen Huang, Ruoyu
  Feng, Jianqiang Huang, Zhibo Chen, and Xian-Sheng Hua.
\newblock Cloth-changing person re-identification from a single image with gait
  prediction and regularization.
\newblock In {\em CVPR}, pages 14278--14287, 2022.

\bibitem{kingma2015adam}
Diederik~P Kingma and Jimmy Ba.
\newblock Adam: A method for stochastic optimization.
\newblock In {\em ICLR}, 2015.

\bibitem{li2019multi}
Jianing Li, Shiliang Zhang, and Tiejun Huang.
\newblock Multi-scale 3d convolution network for video based person
  re-identification.
\newblock In {\em AAAI}, volume~33, pages 8618--8625, 2019.

\bibitem{li2014deepreid}
Wei Li, Rui Zhao, Tong Xiao, and Xiaogang Wang.
\newblock Deepreid: Deep filter pairing neural network for person
  re-identification.
\newblock In {\em CVPR}, pages 152--159, 2014.

\bibitem{li2021learning}
Yu-Jhe Li, Xinshuo Weng, and Kris~M Kitani.
\newblock Learning shape representations for person re-identification under
  clothing change.
\newblock In {\em WACV}, pages 2432--2441, 2021.

\bibitem{lin2021gait}
Beibei Lin, Shunli Zhang, and Xin Yu.
\newblock Gait recognition via effective global-local feature representation
  and local temporal aggregation.
\newblock In {\em ICCV}, pages 14648--14656, 2021.

\bibitem{liu2019spatial}
Yiheng Liu, Zhenxun Yuan, Wengang Zhou, and Houqiang Li.
\newblock Spatial and temporal mutual promotion for video-based person
  re-identification.
\newblock In {\em AAAI}, volume~33, pages 8786--8793, 2019.

\bibitem{lu2022long}
Xiaoyan Lu, Xinde Li, Weijie Sheng, and Shuzhi~Sam Ge.
\newblock Long-term person re-identification based on appearance and gait
  feature fusion under covariate changes.
\newblock {\em Processes}, 10(4):770, 2022.

\bibitem{mccormac2017scenenet}
John McCormac, Ankur Handa, Stefan Leutenegger, and Andrew~J Davison.
\newblock Scenenet rgb-d: Can 5m synthetic images beat generic imagenet
  pre-training on indoor segmentation?
\newblock In {\em ICCV}, pages 2678--2687, 2017.

\bibitem{mclaughlin2016recurrent}
Niall McLaughlin, Jesus~Martinez Del~Rincon, and Paul Miller.
\newblock Recurrent convolutional network for video-based person
  re-identification.
\newblock In {\em CVPR}, pages 1325--1334, 2016.

\bibitem{qian2020long}
Xuelin Qian, Wenxuan Wang, Li Zhang, Fangrui Zhu, Yanwei Fu, Tao Xiang, Yu-Gang
  Jiang, and Xiangyang Xue.
\newblock Long-term cloth-changing person re-identification.
\newblock In {\em ACCV}, 2020.

\bibitem{ros2016synthia}
German Ros, Laura Sellart, Joanna Materzynska, David Vazquez, and Antonio~M
  Lopez.
\newblock The synthia dataset: A large collection of synthetic images for
  semantic segmentation of urban scenes.
\newblock In {\em CVPR}, pages 3234--3243, 2016.

\bibitem{shi2019two}
Lei Shi, Yifan Zhang, Jian Cheng, and Hanqing Lu.
\newblock Two-stream adaptive graph convolutional networks for skeleton-based
  action recognition.
\newblock In {\em CVPR}, pages 12026--12035, 2019.

\bibitem{shu2021semantic}
Xiujun Shu, Ge Li, Xiao Wang, Weijian Ruan, and Qi Tian.
\newblock Semantic-guided pixel sampling for cloth-changing person
  re-identification.
\newblock {\em IEEE Sign. Process. Letters}, 28:1365--1369, 2021.

\bibitem{sun2019dissecting}
Xiaoxiao Sun and Liang Zheng.
\newblock Dissecting person re-identification from the viewpoint of viewpoint.
\newblock In {\em CVPR}, pages 608--617, 2019.

\bibitem{varol2021synthetic}
G{\"u}l Varol, Ivan Laptev, Cordelia Schmid, and Andrew Zisserman.
\newblock Synthetic humans for action recognition from unseen viewpoints.
\newblock {\em IJCV}, 129(7):2264--2287, 2021.

\bibitem{wan2020person}
Fangbin Wan, Yang Wu, Xuelin Qian, Yixiong Chen, and Yanwei Fu.
\newblock When person re-identification meets changing clothes.
\newblock In {\em CVPRW}, pages 830--831, 2020.

\bibitem{wang2022frame}
Likai Wang, Jinyan Chen, and Yuxin Liu.
\newblock Frame-level refinement networks for skeleton-based gait recognition.
\newblock {\em CVIU}, 222:103500, 2022.

\bibitem{welling2016semi}
Max Welling and Thomas~N Kipf.
\newblock Semi-supervised classification with graph convolutional networks.
\newblock In {\em ICLR}, 2017.

\bibitem{xie2022improving}
Qiaokang Xie, Zhenbo Lu, Wengang Zhou, and Houqiang Li.
\newblock Improving person re-identification with multi-cue similarity
  embedding and propagation.
\newblock {\em IEEE TMM}, 2022.

\bibitem{yang2019person}
Qize Yang, Ancong Wu, and Wei-Shi Zheng.
\newblock Person re-identification by contour sketch under moderate clothing
  change.
\newblock 43(6):2029--2046, 2019.

\bibitem{yu2020cocas}
Shijie Yu, Shihua Li, Dapeng Chen, Rui Zhao, Junjie Yan, and Yu Qiao.
\newblock Cocas: A large-scale clothes changing person dataset for
  re-identification.
\newblock In {\em CVPR}, pages 3400--3409, 2020.

\bibitem{zhang2018long}
Peng Zhang, Qiang Wu, Jingsong Xu, and Jian Zhang.
\newblock Long-term person re-identification using true motion from videos.
\newblock In {\em WACV}, pages 494--502, 2018.

\bibitem{zhang2020learning}
Peng Zhang, Jingsong Xu, Qiang Wu, Yan Huang, and Xianye Ben.
\newblock Learning spatial-temporal representations over walking tracklet for
  long-term person re-identification in the wild.
\newblock {\em IEEE TMM}, 23:3562--3576, 2020.

\bibitem{zhang2022bytetrack}
Yifu Zhang, Peize Sun, Yi Jiang, Dongdong Yu, Fucheng Weng, Zehuan Yuan, Ping
  Luo, Wenyu Liu, and Xinggang Wang.
\newblock Bytetrack: Multi-object tracking by associating every detection box.
\newblock In {\em ECCV}, pages 1--21. Springer, 2022.

\bibitem{zhang2019gait}
Ziyuan Zhang, Luan Tran, Xi Yin, Yousef Atoum, Xiaoming Liu, Jian Wan, and
  Nanxin Wang.
\newblock Gait recognition via disentangled representation learning.
\newblock In {\em CVPR}, pages 4710--4719, 2019.

\bibitem{zheng2016mars}
Liang Zheng, Zhi Bie, Yifan Sun, Jingdong Wang, Chi Su, Shengjin Wang, and Qi
  Tian.
\newblock Mars: A video benchmark for large-scale person re-identification.
\newblock In {\em ECCV}, pages 868--884, 2016.

\bibitem{zheng2015scalable}
Liang Zheng, Liyue Shen, Lu Tian, Shengjin Wang, Jingdong Wang, and Qi Tian.
\newblock Scalable person re-identification: A benchmark.
\newblock In {\em ICCV}, pages 1116--1124, 2015.

\bibitem{zheng2017unlabeled}
Zhedong Zheng, Liang Zheng, and Yi Yang.
\newblock Unlabeled samples generated by gan improve the person
  re-identification baseline in vitro.
\newblock In {\em ICCV}, pages 3754--3762, 2017.

\bibitem{zhu2021gait}
Zheng Zhu, Xianda Guo, Tian Yang, Junjie Huang, Jiankang Deng, Guan Huang,
  Dalong Du, Jiwen Lu, and Jie Zhou.
\newblock Gait recognition in the wild: A benchmark.
\newblock In {\em ICCV}, pages 14789--14799, 2021.

\end{thebibliography}
}

\end{document}